\documentclass[10pt,twocolumn,letterpaper]{article}

\usepackage{cvpr}
\usepackage{times}
\usepackage{epsfig}
\usepackage{graphicx}
\usepackage{amsmath}
\usepackage{amssymb}
\usepackage{amsmath}
\usepackage{amssymb}
\usepackage{mathtools}
\usepackage{adjustbox}
\usepackage{gensymb}
\usepackage{indentfirst}
\usepackage{multirow}
\usepackage[T1]{fontenc}
\usepackage{mathrsfs}
\usepackage{booktabs}
\usepackage{array}
\usepackage{colortbl}
\usepackage{subfig}

\cvprfinalcopy 

\def\cvprPaperID{****} 

\begin{document}

\title{DeepMeshFlow: Content Adaptive Mesh Deformation for Robust Image Registration}

\author{Nianjin Ye$^{1,2}$ \hspace{-1.0em}
\and Chuan Wang$^2$ \hspace{-1.0em}
\and Shuaicheng Liu$^{1,2}$ \hspace{-1.0em}
\and Lanpeng Jia$^2$ \hspace{-1.0em}
\and Jue Wang$^2$ \hspace{-1.0em}
\and Yongqing Cui$^1$\\
\vspace{.2em}
\and 1. University of Electronic Science and Technology of China.\\
2. Megvii Technology \\
}


\maketitle

\begin{abstract}
   Image alignment by mesh warps, such as meshflow, is a fundamental task which has been widely applied in various vision applications(e.g., multi-frame HDR/denoising, video stabilization). Traditional mesh warp methods detect and match image features, where the quality of alignment highly depends on the quality of image features. However, the image features are not robust in occurrence of low-texture and low-light scenes. Deep homography methods, on the other hand, are free from such problem by learning deep features for robust performance. However, a homography is limited to plane motions. In this work, we present a deep meshflow motion model, which takes two images as input and output a sparse motion field with motions located at mesh vertexes. The deep meshflow enjoys the merics of meshflow that can describe nonlinear motions while also shares advantage of deep homography that is robust against challenging textureless scenarios. In particular, a new unsupervised network structure is presented with content-adaptive capability. On one hand, the image content that cannot be aligned under mesh representation are rejected by our learned mask, similar to the RANSAC procedure. On the other hand, we learn multiple mesh resolutions, combining to a non-uniform mesh division. Moreover, a comprehensive dataset is presented, covering various scenes for training and testing. The comparison between both traditional mesh warp methods and deep based methods show the effectiveness of our deep meshflow motion model.
\end{abstract}

\section{Introduction}\label{sec:intro}The problem of image registration is a classic vision topic that has been studied for decades~\cite{capel2004image,szeliski2007image,hartley2003multiple}, but still active not only because its difficulty under certain circumstances, such as low-textured scenes, parallax and dynamic objects, but also due to its widespread applications, such as Panorama creation~\cite{brown2003recognising}, multi-frame HDR/Denoising~\cite{zhang2010denoising}, multi-frame super resolution~\cite{Wronski2019}, and video stabilization~\cite{liu2013bundled}.

\begin{figure}[t]
\begin{center}
\includegraphics[width=1.0\linewidth]{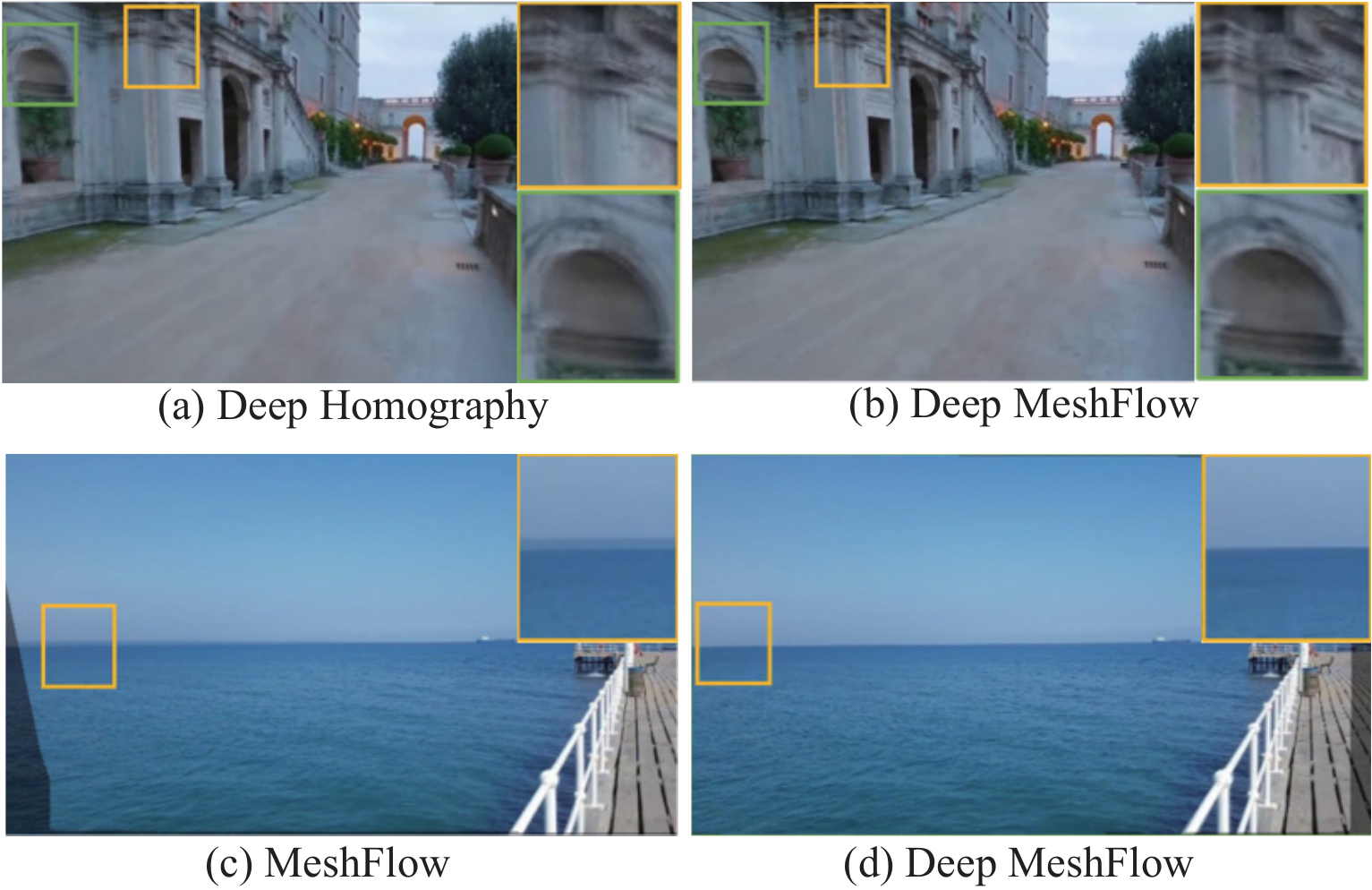}
\end{center}
\vspace{-0.5em}
   \caption{Comparison between popular image registration methods, deep homography~\cite{nguyen2018unsupervised}, Meshflow~\cite{liu2016meshflow} and our proposed Deep Meshflow. The source images are aligned to the target image. Two images are blended. Good registration produces less blurs and is free from ghosting effects. We show zoom-in windows to highlight misalignments. }
\vspace{-0.5em}
\label{fig:teaser}
\end{figure}

A variety of motion models have been proposed for image registration, among which the homography~\cite{hartley2003multiple} is the most popular one given its simplicity and efficiency. Homography is estimated by matching image features~\cite{lowe2004distinctive} between two images. The false matches are rejected by RANSAC~\cite{fischler1981random}. One problem is that the quality of estimated homography is highly dependent on the quality of matched image features. Insufficient number of correct matches or uneven distributions can easily damage the performance. Recently, deep homography has been proposed which takes two images as input to the network and output the homography~\cite{detone2016deep,nguyen2018unsupervised}. Compared with feature-based methods, deep homography is more robust against various challenging cases, such as low-light, low-texture, high-noise, etc~\cite{zhang2019content}. The other problem of homography is its limited degree of freedom. A homography can only describe plan motions or motions caused by camera rotations. Violation of these assumptions can produce incorrect alignments. For images with parallax, a global homograhy is usually used to estimate an initial alignment before subsequent more sophisticated models~\cite{gao2011constructing,liu2009content,zaragoza2013projective,lin2016seagull}.
Mesh-based image warping can represent spatially varying depth variations~\cite{liu2009content,liu2013bundled}. Each of mesh grid undergoes a local linear homography, accumulating to a highly nonlinear representation. Igarashi~\emph{et al.} proposed as-rigid-as-possible image warping to enforce local rigidity of each mesh triangles~\cite{igarashi2005rigid}. Later, Liu~\emph{et al.} extends \cite{igarashi2005rigid} by proposing content-preserving warps that constrains the rigidity of mesh cells according to the image contents~\cite{liu2009content}. Liu~\etal. proposed a meshflow motion model that further simplify the estimation of mesh model~\cite{liu2016meshflow}. The meshflow contains a sparse motion field with motions only located at the mesh vertexes. These methods have been proven to be sufficiently flexible for handling complex scene depth variations~\cite{li2015dual,liu2014fast,liu2013bundled}. However, one common challenge faced by mesh-based methods is still the quality of image features. Both number of matched features and feature distribution can influence the performance. Zaragoza~\emph{et al.}~\cite{zaragoza2013projective} proposed an As-Projective-As-Possible (APAP) mesh deformation approach and Lin~\emph{et al.}~\cite{Lin2011Smoothly} proposed a spatially varying affine model to alleviate the problem of feature dependent by interpolating ideal features to non-ideal regions. However, they still require a certain number of qualified features to start with. Optical flow~\cite{Sun2010Secrets}, on the other hand, can estimate per-pixel motions which can preserve fine motion details compared with mesh interpolated motion fields. However, optical flow estimation is more computationally expensive compared with light-weight mesh representations. Synthesis driven applications do not require physically accurate motion estimation at every pixel, estimating optical flow over-kills the requirement and is often not necessary.

Figure~\ref{fig:teaser} shows some examples. Figure~\ref{fig:teaser}(a) and (b) show the comparison between our method and deep homography method~\cite{nguyen2018unsupervised}. The source image is aligned to the target image and two images are blended for illustration. The scene contains multiple planes, e.g., ground and building facade. Misalignments (highlighted by the zoom-in window) can be observed at the building facade of deep homography method while our deep meshflow can align multiple planes and thus, is free from such problem. Figure~\ref{fig:teaser}(c) and (d) show the comparison of the traditional meshflow~\cite{liu2016meshflow}. The feature detection is difficult in this example due to the poor textures, causing the failure of meshflow. In contrast, our deep meshflow is robust to textureless scenes.

In this work, we propose an unsupervised approach with a new architecture for content adaptive deep meshflow estimation. We combine the advantage of deep homography that is robust to textureless regions, and the advantage of meshflow that is light weight for nonlinear motion representation. Specifically, our network takes two images for alignment as input, and output a sparse motion field meshflow, with motions only located at mesh vertexes. We learn a content mask to reject outlier regions, such as moving objects and discontinues large foreground that cannot be registed by the mesh deformation but can influence the overall alignment quality. The capability of content adaptive is similar as the RANSAC procedure when estimating homography~\cite{hartley2003multiple} or mesh warps~\cite{liu2013bundled,liu2016meshflow} in traditional approaches. This is realized by a novel triplet loss. Moreover, instead of directly output the mesh at the desired resolution, we first generate several intermediate meshes with different resolutions, e.g., meshes with $1\times1$, $4\times4$, $16\times16$ etc. Then we choose the best combination among these meshes, assembling to the final output. This idea is borrowed from video coding x265~\cite{sullivan2012overview}, in which the block division of a frame can be non-uniform according to the image content. Here, our mesh division is also non-uniform based on both image contents and motions. For regions that require higher degree of freedom, we chose finer scales for registration accuracy, while for regions that are relatively complanate, we chose coarse scales for robustness. This flexibility is realized by our segmentation module in the pipeline, which shows to be more effective than simply chose the finest scale.

In addition, we introduce a comprehensive meshflow dataset for training, within which the testing set contains manually labeled ground-truth point matches for the purpose of evaluation. We split the dataset into $5$ categories according to the scene characteristics, including scenes with multiple dominate planes, scenes captured at night, scenes with low-textured regions, with small-foreground and with large-foreground. The experiments show that our method outperforms previous leading traditional mesh-based methods~\cite{liu2016meshflow,zaragoza2013projective}, as well as recent deep homography methods~\cite{detone2016deep,nguyen2018unsupervised,zhang2019content}. Our contributions can be summarized as:
\begin{itemize}
  \item A new unsupervised network structure for deep meshflow estimation, which outperforms previous state-of-the-arts methods.
  \item The content-adaptive capability, in terms of rejecting interference regions and adaptive mesh scale selection.
  \item A comprehensive dataset contains various scene types for training and testing.
\end{itemize}

\begin{figure*}[t!]
  \centering
  \subfloat[][Network structure]{
  \includegraphics[width=0.68\linewidth]{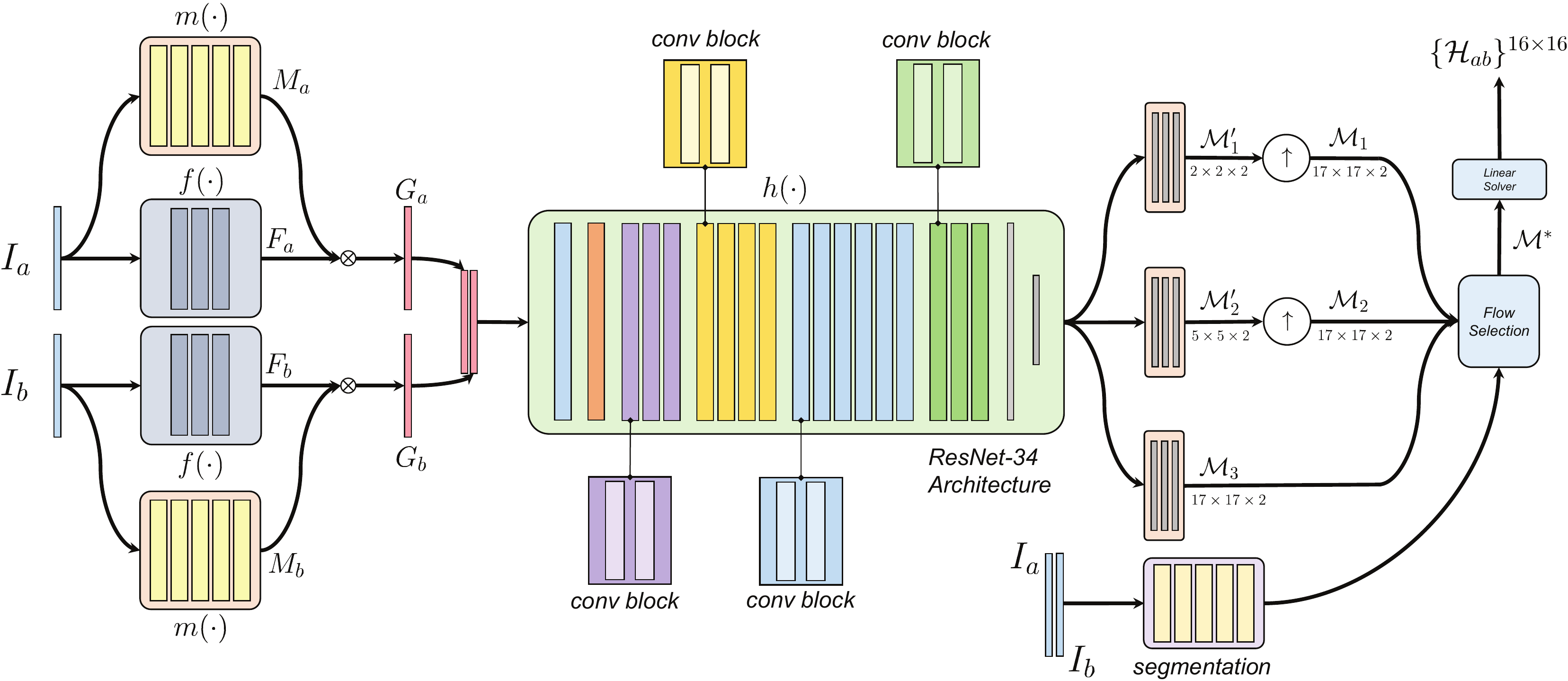}}
  \hfill
  \subfloat[][Triplet loss]{
  \includegraphics[width=0.29\linewidth]{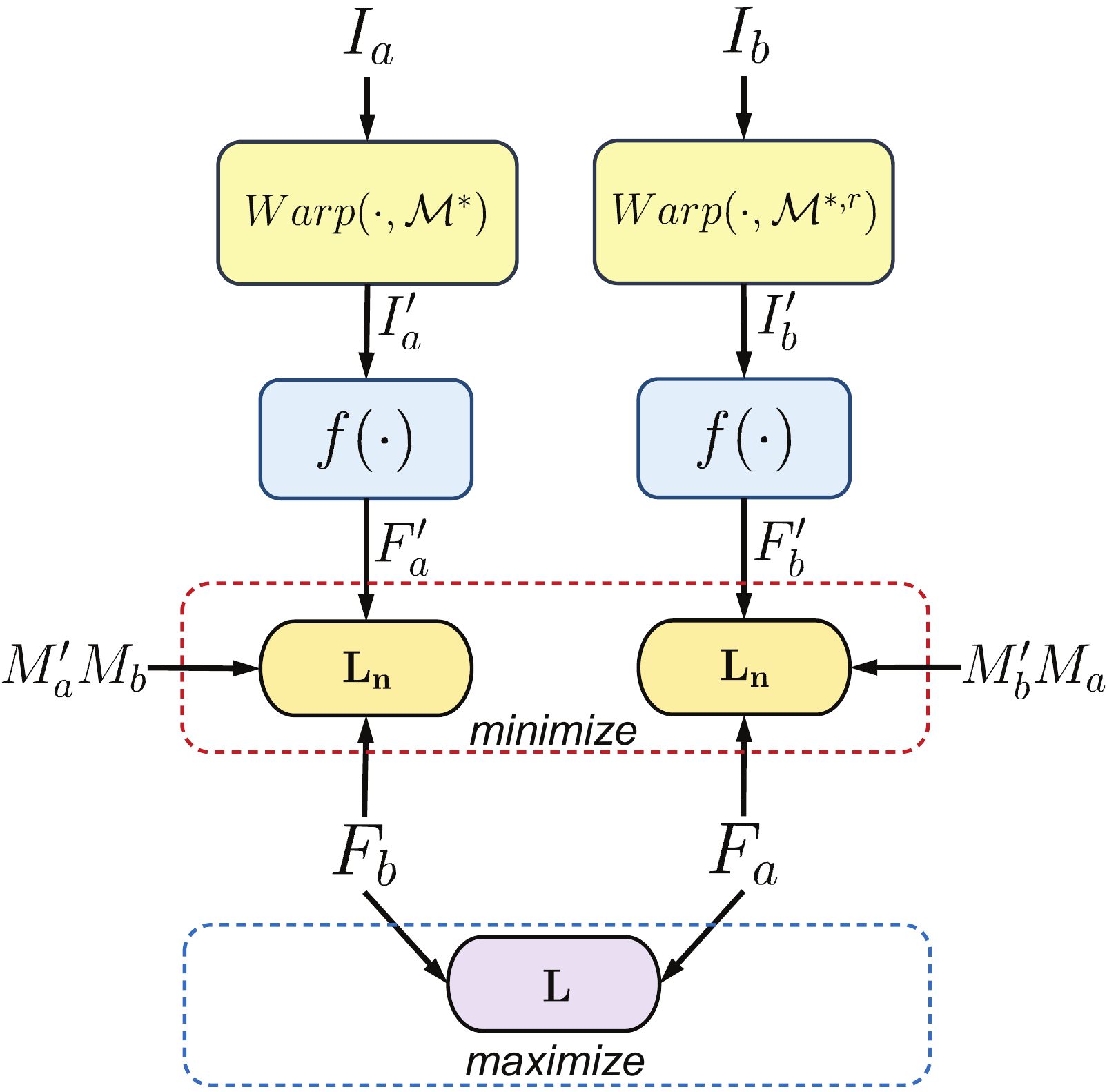}}
  \caption{Network structure (a) and triplet loss (b) used for our DeepMeshFlow estimation.}\label{fig:pipeline}
\end{figure*}

\section{Related Works}\label{sec:related_works}
\paragraph{Global parametric models.}
Homography is a wildly used parametric alignment model, which is a $3 \times 3$ matrix with $8$ degree of freedom, describing either plan motions in the space or motions induced by pure camera rotations. Traditional methods require sparse feature matches~\cite{lowe2004distinctive,bay2006surf,rublee2011orb} to estimate a homography. However, image features are unreliable with respect to low-textured regions. Recently, deep based solutions have been proposed for improved robustness such as, the supervised approach that train homographyNet under the guidance of random homography proposals~\cite{detone2016deep} or unsupervised approach that directly minimizes warping MSE distance~\cite{nguyen2018unsupervised}. On the other hand, homography model is restricted by its motion assumptions, violation which can easily introduce misalignments, such as scenes consisting of multiple plans or discontinuity depth variations.

\paragraph{Mesh warping.}
To solve the depth parallax issue, mesh-based image warping is more popular. Liu~\emph{et al} proposed Content Preserving Warp (CPW) to encourage mesh cells to under go a rigid motion~\cite{liu2009content}.  Li~\emph{et al.} proposed a duel-feature warping by considering not only image features but also line segments for the warping in low-textured regions~\cite{li2015dual}. Lin~\emph{et al.} incorporated curve preserving term to preserve curve structures~\cite{lin2016seamless}. Liu~\emph{et al.} introduced MeshFlow, a non-parametric warping method for video stabilization~\cite{liu2016meshflow}. Compared with dense optical flow, meshflow is a sparse motion field with motions only located at mesh vertexes. It detects and tracks image features for meshflow model estimation. In this work, we propose a deep solution, DeepMeshFlow, for the similar purpose, but with largely improved robustness against scenes that suffer from feature detection and matching/tracking problems.

\paragraph{Optical flows.}
Optical flow estimates per-pixel dense motion between two images. Compared with global alignment methods, optical flow can produce better results in preserving motion details. The traditional method often adopt coarse-to-fine, variational optimization framework for flow estimation~\cite{Horn1981,Sun2010Secrets,Wulff2015Efficient}. Recently, flow accuracy has been largely promoted by convolutional networks~\cite{Philippe2013DeepFlow,Sun2018PWC,Ilg2017FlownetV2}. For some image/video editing applications, however, the optical flow often requires a series of post-processing before the usage, such as occlusion detection, motion inpainting, outlier filtering. For example, Liu \etal, estimated a Steadyflow from raw optical flow by rejecting and inpainting motion inconsistent foregrounds. Our mesh-based representation, on the other hand, is free from such issues. It is light-weight and flexible for various applications, such as multi-frame HDR~\cite{zhang2010denoising}, burst denoising~\cite{liu2014fast}, and video stabilization~\cite{liu2009content,liu2013bundled,liu2016meshflow}.

\section{Algorithm}\label{sec:algo}
MeshFlow is a motion model that describes non-linear warping between two image views~\cite{liu2016meshflow}. It has more degrees of freedom compared with homography but suffers less from computational complexity compared with optical flow. It is represented by a mesh of $H_g\times W_g$ grids so that totally contains $(H_g+1)\times(W_g+1)$ vertices on the mesh. At each vertex, a 2D motion vector is defined so that each grid corresponds to one homography computed by the 4 vectors on its 4 corner vertices. With multiple homography matrices computed on the various mesh grids, the entire image can be warpped in an non-linear manner so as to fit multi-planes in the scene.

\subsection{Network Structure}
Our method is built upon convolutional neural network which takes two images $I_a$ and $I_b$ as input, and produces a mesh flow $\mathcal{M}^*$ of size $(H_g + 1)\times(W_g + 1)\times2$ as output, where $H_g$ and $W_g$ are the height and width of the mesh with a 2D motion vector being defined on each vertex of the mesh. Given the mesh flow with such a form, each grid of it can be represented by a homography matrix $\mathcal{H}_{ab}$, solved by the 4 motion vectors on its 4 corners. The entire network structure can be divided into four modules: a feature extractor $f(\cdot)$, a mask predictor $m(\cdot)$, a scene segmentation network $s(\cdot)$ and a multi-scale mesh flow estimator $h(\cdot)$. $f(\cdot)$ and $m(\cdot)$ are fully convolutional networks which accept input of arbitrary sizes and produce a concatenation of feature maps. Then $h(\cdot)$ servers as a regressor that transfers the features into $K$ mesh flows in multiple scales. Then, a scene segmentation network $s(\cdot)$ produces a fusion mask that fuses the multi-scale mesh flows into one as the final output. Figure~\ref{fig:pipeline} illustrates the network structure, and in this sub-section we briefly introduce $f(\cdot), m(\cdot)$ and $s(\cdot)$ and leave $h(\cdot)$ into the next sub-section.

\vspace{-0.5em}
\paragraph{Feature extractor.}
Unlike previous DNN based methods that directly utilizes the pixel values as the feature, here our network automatically learns a feature from the input for robust feature alignment. To this end, we build a FCN that takes an input of size $H\times W\times 3$, and produces a feature map of size $H\times W\times C$. For inputs $I_a$ and $I_b$, the feature extractor shares weights and produces feature maps $F_a$ and $F_b$, i.e.
\begin{equation}
    F_a = f(I_a), \quad F_b = f(I_b)
\end{equation}

\paragraph{Mask predictor.}
In non-planar scenes, especially those including moving objects, there exists no single homography that can align the two views. Although mesh flow contains multiple homography matrices which can partially solve the non-planar issue, for a local single region, one homography could be still problematic to well align all the pixels. In traditional algorithm, RANSAC is widely applied to find the inliers for homography estimation, so as to solve the most approximate matrix for the scene alignment. Following the similar idea, we build a sub-network to automatically learn the inliers' positions. Specifically, a sub-network $m(\cdot)$ learns to produce an inlier probability map or mask, highlighting the content in the feature maps that contribute much for the homography estimation. The size of the mask is the same as the size of the feature. With the masks, we further weight the features extracted by $f$ before feeding them to the homography estimator, obtaining two weighted feature maps $G_a$ and $G_b$ as,
\begin{alignat}{3}
&M_a &&= m(I_a),\quad &G_a &= F_a M_a \\
&M_b &&= m(I_b),\quad &G_b &= F_b M_b
\end{alignat}

\paragraph{Scene segmentation.}
The weighted feature maps $G_a$ and $G_b$ are concatenated and fed to the following MeshFlow estimator $h(\cdot)$, to produce $K$ mesh flows with different scales. These multi-scale mesh flows are then fused into one by a branch-selection scheme. It is achieved by training a scene segmentation network that segments the image $I$ into $K$ classes, each one of which corresponds to one branch, i.e.
\begin{align}
    S = s(I_a, I_b)
\end{align}
where $S$ is of the same resolution of the finest-scale mesh flow, so its size is $(H_g+1)\times(W_g+1)\times K$.

\begin{figure*}[t]
\centering
\includegraphics[width=1.0\linewidth]{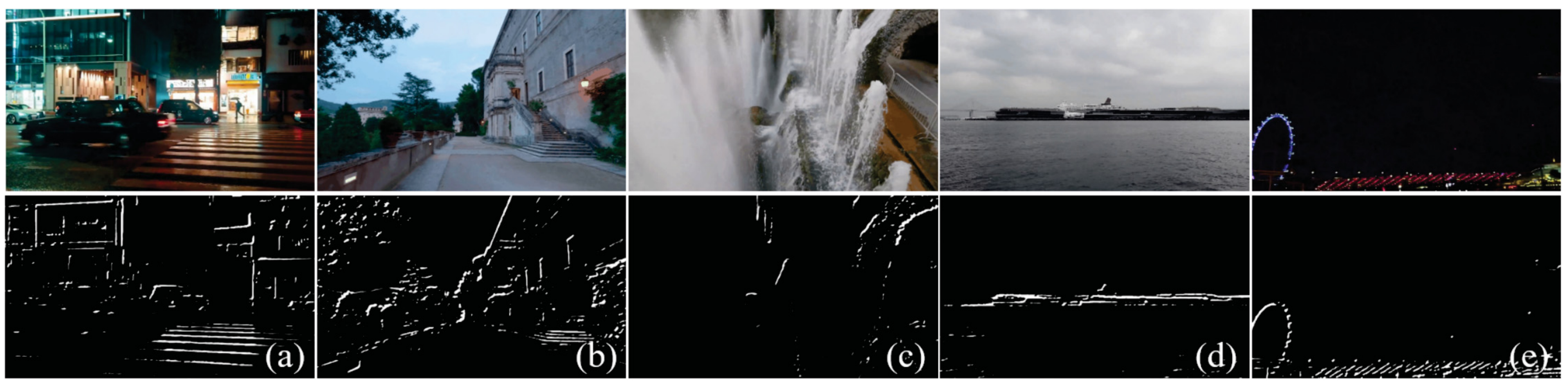}
   \caption{Our predicted mask for various of scenes. (a) contains a moving car which is removed in the mask. (b) is a regular static scene that the mask equally looks at the entire image. (c) contains large foreground, the fountain, which is successfully rejected by the mask. (d) consists of sea and sky, which cannot provide rich textures. The mask concentrate on the horizon. (e) is an night example. The mask looks at the sky-wheel and buildings. }
\label{fig:mask}
\end{figure*}

\subsection{MeshFlow Estimator}
\paragraph{Multi-scale MeshFlow.}
As mentioned above, the output of our network is a mesh flow $\mathcal{M}^*$ of size $(H_g + 1)\times(W_g + 1)\times 2$. Directly regressing the input, i.e. the two weighted feature maps $G_a$ and $G_b$ to this mesh flow is not straightforward, as there exists too many degrees of freedom (DoF) being involved. To tackle this issue, we divide the mesh flow regression part into $K$ branches, each of which is responsible for one scale mesh flow. The intuition behind results from the fact that in complex scenes, various planes may differ in scales. A coarse-scaled mesh flow could be better align the two views rigidly, and trends to be easier for training compared with a fine-scaled mesh flow of more DoF. As for its backbone, it follows a ResNet-34 structure, which contains 34 layers of strided convolutions followed by $K$ branches, each of which starts with an adaptive pooling layer and generates a mesh flow with specific size by an additional convolutional layer. In our experiments, we set $K$ to 3 so that the 3 branches correspond to mesh flow $\mathcal{M}'_1$ of size $(\frac{H_g}{16} + 1)\times(\frac{W_g}{16} + 1)\times2$, $\mathcal{M}'_2$ of size $(\frac{H_g}{4} + 1)\times(\frac{W_g}{4} + 1)\times2$ and $\mathcal{M}_3$ of size $(H_g + 1)\times(W_g + 1)\times 2$. The coarse-scaled mesh flows $\mathcal{M}'_1, \mathcal{M}'_2$ are then upsampled to the same scale of $\mathcal{M}_3$ before fusing together, noted as $\mathcal{M}_1, \mathcal{M}_2$. This process is expressed as follows,
\begin{align}
    \{\mathcal{M}_k\} = h(G_a, G_b), \quad k = 1,2,...,K
\end{align}

\paragraph{MeshFlow fusion.}
With $\{\mathcal{M}_k\}$ computed by the previous steps, we finally fuse the $K$ mesh flows into the output mesh flow using the segmentation mask $S$ in the following manner,
\begin{align}
\mathcal{M}^*(u, v) = \mathcal{M}_{\tilde{k}}(u, v)
\end{align}
where $\tilde{k} = \arg\max_k S(u,v,k)$ and $(u, v)$ is the vertex coordinate on the mesh. By this strategy, the output mesh flow conveys homography alignment in various scales for each local grid. It has enough DoF to align the two views and is still easy for training.

\subsection{Triplet Loss for Training}
With the mesh flow $\mathcal{M}^*$ estimated, we obtain $\{\mathcal{H}_{ab}\}^{16\times16}$ by computing the homography matrix for each of its grid. Then we warp image $I_a$ to $I'_a$ and then further extracts its feature map as $F'_a$. Intuitively, for a local grid $(u,v)$, if the homography matrix $\mathcal{H}_{ab}(u,v)$ is accurate enough, $F'_a$ should be well aligned with $F_b$, causing a low $l_1$ loss between them. Considering in real scenes, a single homography matrix cannot satisfy the transformation between the two views, we also normalize the $l_1$ loss by $M'_a$ and $M_b$. Here $M'_a$ is the warped version of $M_a$. So the loss between the warped $I_a$ and $I_b$ is as follows,
\begin{align}
\mathbf{L_n}(I'_a, I_b) = \frac{\sum_i M'_aM_b\cdot|F'_a - F_b|}{\sum_i M'_aM_b} \label{eq:l1-warp-Ia-Ib}
\end{align}
where $F'_a = f(I'_a),~~I'_a = Warp(I_a, \mathcal{M}^*)$ and $i$ indicates a pixel location in the masks and feature maps. Here we utilize spatial transform network~\cite{jaderberg2015spatial} to achieve the warping operation.

Directly minimizing Eq.~\ref{eq:l1-warp-Ia-Ib} may easily cause trivial solutions, where the feature extractor only produces all zero maps, i.e. $F'_a = F_b = 0$. In this case, the features learned indeed describe the fact that $I'_a$ and $I_b$ are well aligned, but it fails to reflect the fact that the original images $I_a$ and $I_b$ are mis-aligned. To this end, we involve another loss between $F_a$ and $F_b$, i.e.
\begin{align}
\mathbf{L}(I_a, I_b) = |F_a - F_b| \label{eq:l1-Ia-Ib}
\end{align}
and further maximize it when minimizing Eq.~\ref{eq:l1-warp-Ia-Ib}. This strategy avoids the trivial solutions, and enables the network to learn a discriminative feature map for image alignment.

In practise, we swap the features of $I_a$ and $I_b$ and produce another reversed mesh flow $\mathcal{M}^{*,r}$, and a homography matrix $\mathcal{H}_{ba}(u,v)$ is computed for each grid. Following Eq.~\ref{eq:l1-warp-Ia-Ib} we involve a loss $L(I'_b,I_a)$ between the warped $I_b$ and $I_a$. We also add a constraint that enforces $\mathcal{H}_{ab}(u,v)$ and $\mathcal{H}_{ba}(u,v)$ to be inverse. So, the optimization procedure of the network could be written as follows,
\begin{align}
\min_{m, f, h}  &
\mathbf{L_n}(I'_a, I_b) + \mathbf{L_n}(I'_b, I_a) - \lambda \mathbf{L}(I_a, I_b)\nonumber \\
& + \mu \sum_{(u,v)}|\mathcal{H}_{ab}(u,v)\mathcal{H}_{ba}(u,v) - \mathcal{I}|,
\label{eq:tripleloss}
\end{align}
where $\lambda$ and $\mu$ are balancing hyper-parameters, and $\mathcal{I}$ is a 3-order identity matrix. We set $\lambda = 2.0$ and $\mu = 0.01$ in our experiments. We illustrates the loss formulations in \figurename~\ref{fig:pipeline}(b).

\subsection{Unsupervised Content-Awareness Learning}
As mentioned above, our network contains a sub-network $m(\cdot)$ to predict an inlier probability map or mask. It is such designed that our network can be of content-awareness by the two-fold effects. First, we use the masks $M_a, M_b$ to explicitly weight the features $F_a, F_b$, so that only highlighted features could be fully fed into MeshFlow estimator $h(\cdot)$. Meanwhile, they are also implicitly involved into the normalized $l_1$ distance between the warped feature $F'_a$ and its original counterpart $F_b$, or $F'_b$ and $F_a$, meaning only those regions that are really fit for alignment would be taken into account. For those areas containing low texture or moving foreground, because they are non-distinguishable or misleading for alignment, they are naturally removed for local homography estimation in a grid during optimizing the triplet loss as proposed. Such a content-awareness is achieved fully by an unsupervised learning scheme, without any ground-truth mask data as supervision.

\begin{figure}[t]
\centering
\includegraphics[width=1.0\linewidth]{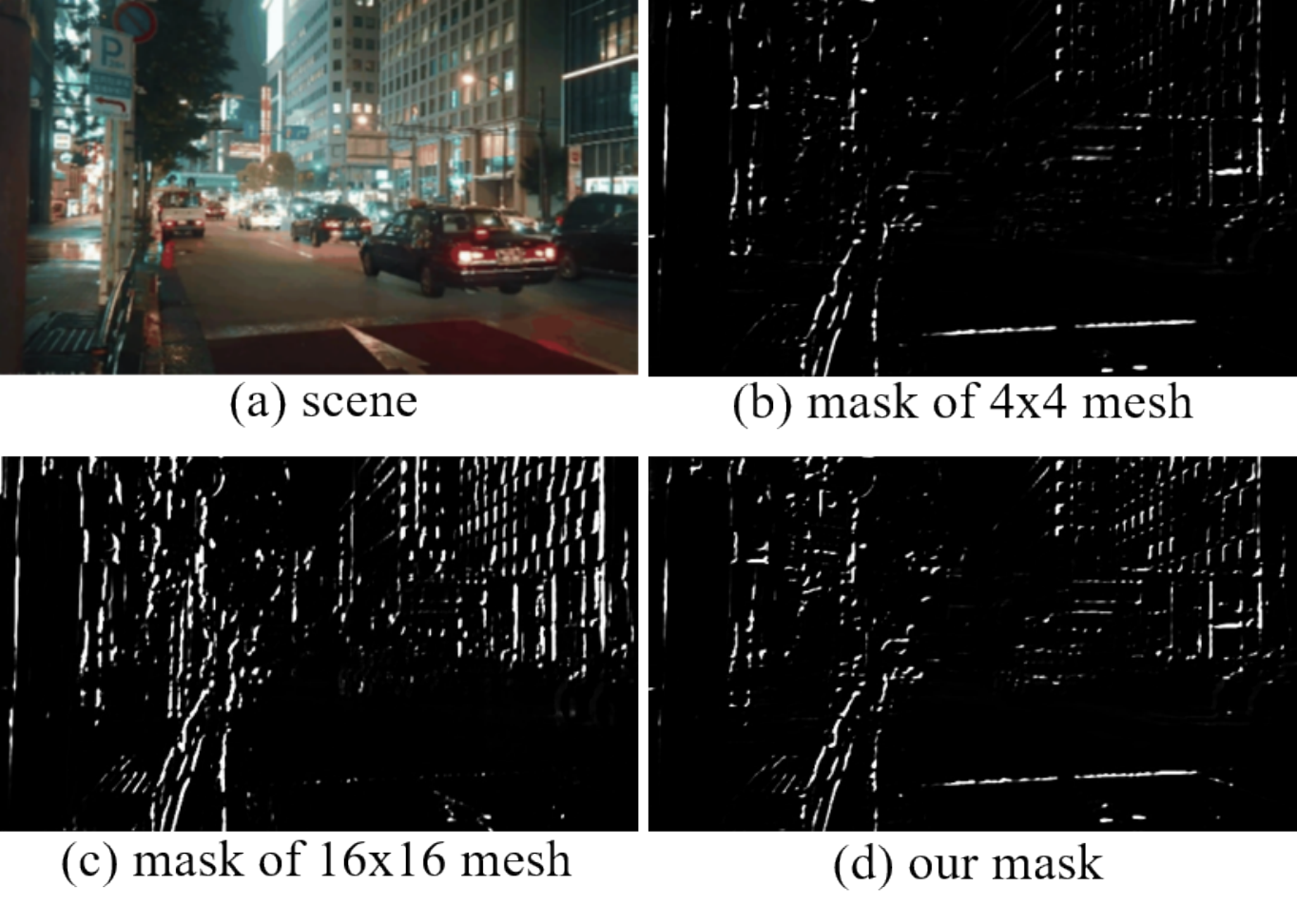}
   \caption{Comparison of masks with respect to different mesh resolutions. (a)the scene; (b)mask produced by $4\times4$ mesh; (c)mask produced by $16\times16$ mesh; and (d)mask produced by our adaptive mesh.}
\label{fig:maskgrad}
\end{figure}

To demonstrate the effectiveness of mask, we illustrate an example in Figure~\ref{fig:maskgrad} and~\ref{fig:mask}. In Figure~\ref{fig:maskgrad}, we visualize the mask if one branch of mesh flow is used only. In this case, for coarse-scaled mesh flow, since each grid covers larger area where a single homography is less likely to represent the transformation, less pixels are highlighted in the mask. However, our DeepMeshFlow solution works in multiple scales, so the highlighted region in the mask is less than the one in mask trained with $16\times 16$ mesh flow, but more than the one in mask trained with $4\times 4$ mesh flow. Figure~\ref{fig:mask} shows mask examples generated in several scenarios. For example, in Figure~\ref{fig:mask}(a)(c) where the scenes contain dynamic objects, our network successfully rejects moving objects, even if the movements are inapparent as the water in (c). These cases are very difficult for RANSAC to find robust inliers. In particular, the most challenging case is Figure~\ref{fig:mask}(a), in which the moving foregrounds are complex, including people and the cars. Our method successfully locates the useful background for the homography estimation. Figure~\ref{fig:mask}(d) is a low-textured example, in which the sky occupies half space of the image. It is challenging for traditional methods where the sky provides no features and the sea causes matching ambiguities. Our predicted mask concentrates on the horizon but with sparse weights on sea waves. Figure~\ref{fig:mask}(e) is a low light example, where only visible areas contain weights as seen. We also conduct an ablation study to reveal the influence if disabling the mask prediction. As seen in Table~\ref{tab:ablation}, the accuracy has a significant decrease when mask is removed.

\section{Experimental Results}\label{sec:exps}

\subsection{Dataset and Implementation Details}
\begin{figure}[t]
\centering
\includegraphics[width=1.0\linewidth]{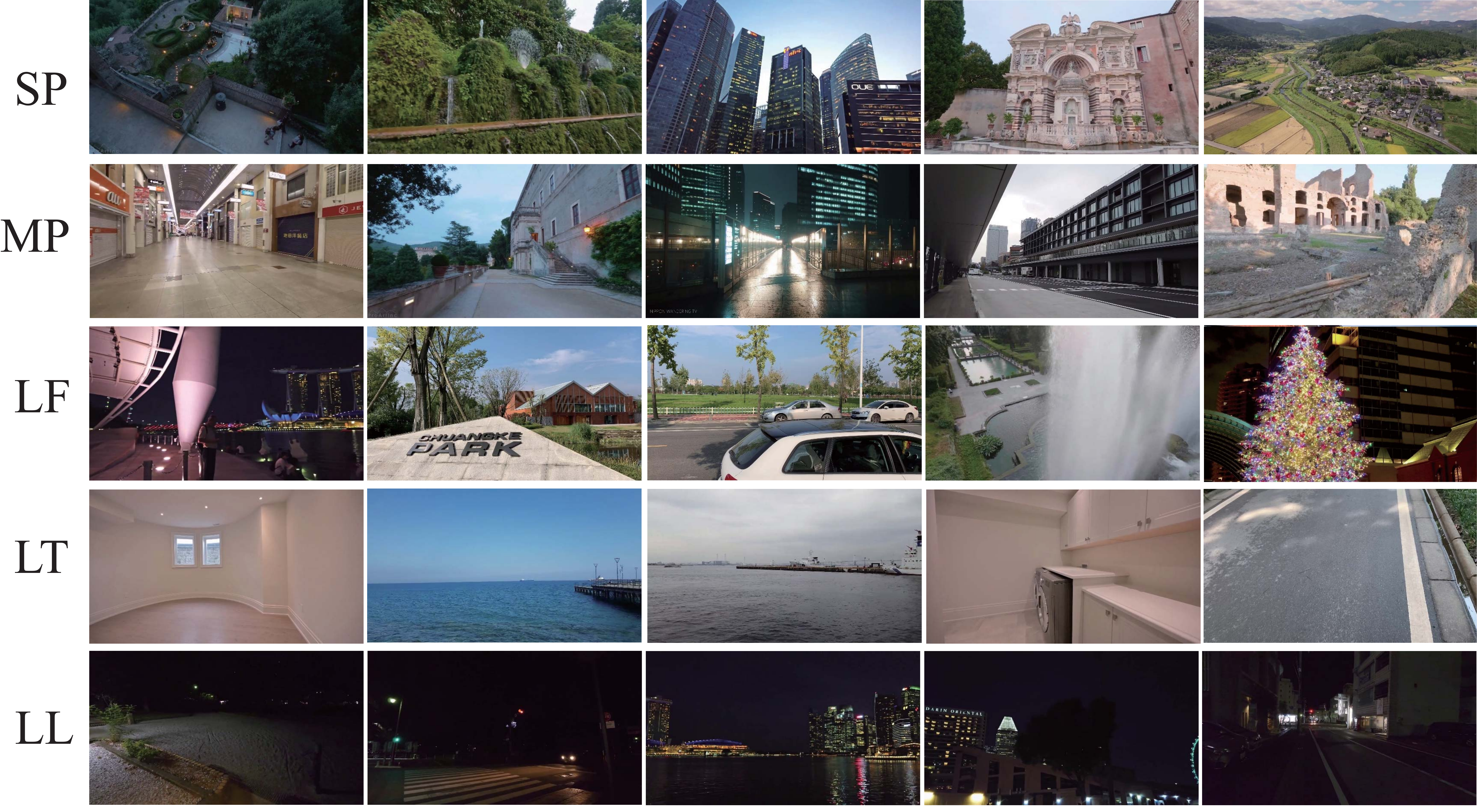}
   \caption{Examples in our dataset. (a) Examples of single plane (SP), (b) examples of multiple plane (MP), (c) examples of scenes contains large foreground (LF), (d) examples of scenes with low-textures (LT) and (e) examples of scenes with low-light (LL). }
\label{fig:dataset}
\end{figure}

To train and evaluate our deep meshflow, we present a comprehensive dataset that contains various of scenes as well as marked point correspondences. We split our dataset into several categories to test the performances under different scenarios. The categories includes: scenes consists of a single plane (SP), scenes mainly consists of multiple dominate planes (MP), scenes with large foreground (LF), scenes with low-textures (LT) and scenes captured with low-light (LL). The first three categories focus on the motion representation capability of motion models while the last two categories concentrate on the capability of feature extraction. Notably, for category LT and LL, they contain all type of scenes SP, MP, and LF. In particular, each category contains around $24k$ image pairs, thus totally $121k$ image pairs in the dataset. Figure~\ref{fig:dataset} shows some examples.

\begin{figure}[t]
\centering
\includegraphics[width=1.0\linewidth]{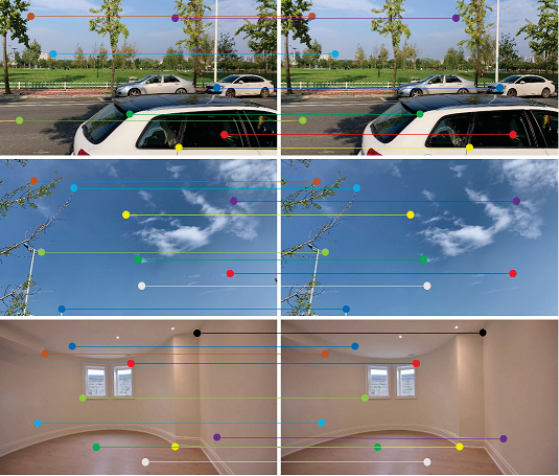}
   \caption{Examples of our annotated point correspondences for quantitative evaluation. The first example contains dominate foreground, we mark points on both foreground and background. The second example films the blue sky, we still mark some points according to textures of clouds. The third example films a textureless indoor white wall, we mark correspondences according to the textures and corners. }
\label{fig:annotation}
\end{figure}

\paragraph{Points annotations.}
For the testing set, we mark ground-truth point correspondence for the purpose of quantitative evaluation. Figure~\ref{fig:annotation} shows several examples of our annotated correspondences. For each pair, we carefully marked around $10$ correspondences which equally distributed on the image. For category multi-plane(MP), we equally separate points on different planes. For category of low-textures(LT), we mark points with extra efforts to make sure the correctness. We marked about 3,000 pairs of images and nearly 30k pairs of matching points for all categories. Figure~\ref{fig:annotation} shows three examples of our annotation.

\begin{figure*}[t]
\centering
\includegraphics[width=1.0\linewidth]{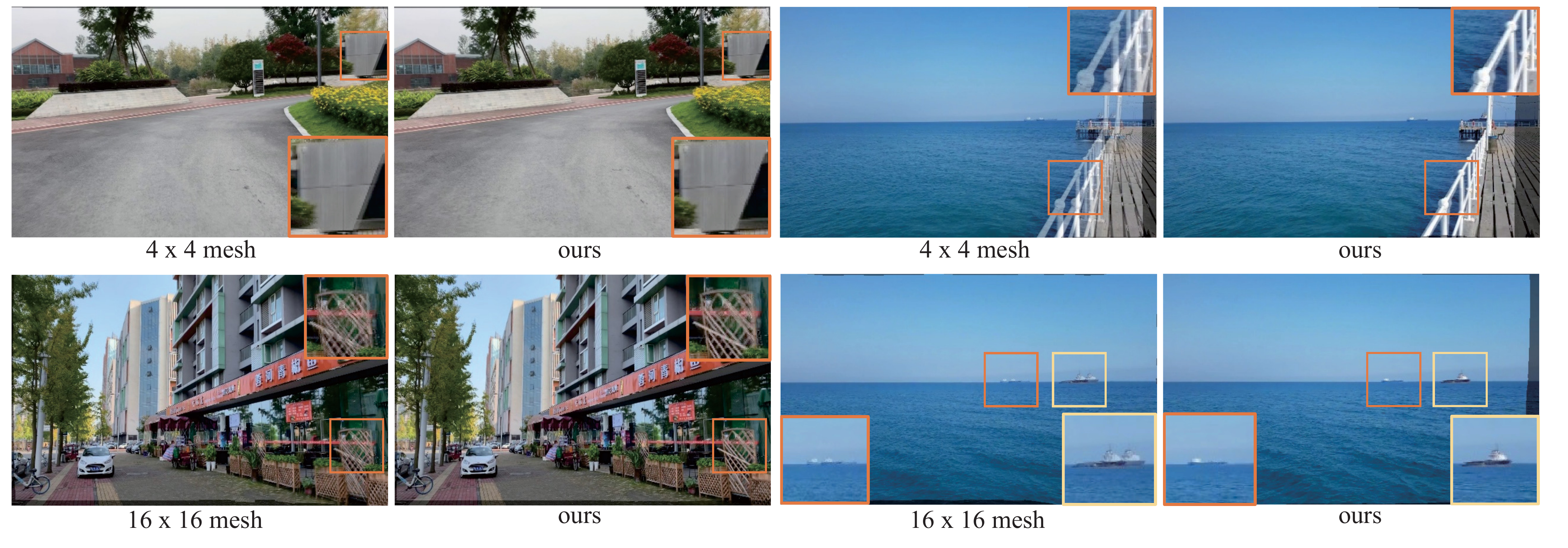}
   \caption{Comparisons with fixed mesh resolutions. We conduct $4\times 4$ and $16\times 16$ fixed meshes to compare with out adaptive mesh. Fixed mesh resolution cannot produce comparable results as ours. For example, the first row, second example, we need a more dense mesh to align nearby handrail, where $4\times 4$ mesh cannot handle this case. On the other hand, the second row, second example, we require a sparse mesh to align faraway ship. Dense mesh may not receive sufficient constraints as nearby regions (blue sky and sea) do not contain rich textures. }
\label{fig:meshresolution}
\end{figure*}

\paragraph{Implementation details.}
Our network is trained with 30k iterations by an Adam optimizer~\cite{kingma2014adam}, whose parameters are set as $l_r = 1.0\times 10^{-4}$,
$beta_1 = 0.9$, $beta_2 = 0.999$, $\varepsilon = 1.0\times 10^{-8}$. The batch size is set to $80$. For every $4k$ iterations, we multiply the learning rate by $0.8$. Each iteration costs approximate $2.1$s and it takes nearly $18$ hours to complete the entire training. The implementation is based on PyTorch and the network training is performed on $8$ NVIDIA
RTX 2080 Ti. To augment the training data and avoid black boundaries appearing in the warped image, we randomly crop patches of size $321\times545$ from the original image to form $I_a$ and $I_b$.

\subsection{Comparison with Existing Methods}
\begin{table}[t!]
  \centering
  \resizebox{1.0\linewidth}{!}{
    \begin{tabular}{>{\arraybackslash}p{2.6cm}
    >{\centering\arraybackslash}p{0.7cm}
    >{\centering\arraybackslash}p{0.7cm}
    >{\centering\arraybackslash}p{0.7cm}
    >{\centering\arraybackslash}p{0.7cm}
    >{\centering\arraybackslash}p{0.7cm}
    >{\centering\arraybackslash}p{0.7cm}}
    \toprule
          & SP    & MP     & LF    & LT    & LL    & Avg \\
    \midrule
    Eye & 6.70  & 8.99  & 4.73  & 7.38  & 7.83  & 7.13 \\
    \midrule
    w/o. Mask & 1.82 & 2.37 & 2.42 & 3.16 & 2.72 & 2.50 \\
    \midrule
    1$\times$1 mesh & 1.78  & 2.24  & 2.31  & 2.40  & 2.72  & 2.29 \\
    4$\times$4 mesh & 1.60  & 1.80  & 2.11  & 2.57  & 2.64  & 2.14 \\
    16$\times$16 mesh & 1.64  & 2.02  & 2.30  & 3.26  & 3.02  & 2.45 \\
    \midrule
    Meshflow & 1.64 & 2.03 & 2.26 & 3.19 & 3.10 & 2.44\\
    Unsupervised & 1.87 & 2.63 & 2.57 & 2.69 & 2.46 & 2.45\\
    \midrule
    Ours & \textbf{1.57} & \textbf{1.74} & \textbf{1.99} & \textbf{2.20} & \textbf{2.45} & \textbf{1.99} \\
    \bottomrule
    \end{tabular}
    }
    \caption{Ablation studies on mask, triple loss, training strategy and network backbones. Data represents the $l_2$ distances between transformed points and marked ground-truth points.}
    \label{tab:ablation}\vspace{-3mm}
\end{table}%

\paragraph{Qualitative comparison.}
We compare our method with various methods, including classic traditional methods MeshFlow~\cite{liu2016meshflow}, As-Projective-As-Possible mesh Warping~\cite{zaragoza2013projective} and a deep method, supervised deep homography~\cite{detone2016deep}. For the unsupervised deep homography method, it uses aerial images as training data, which ignores the effect of depth parallex. For a more fair comparison, we fine-tune the method with our training data.

\begin{figure*}[t]
\centering
   \includegraphics[width=1.0\linewidth]{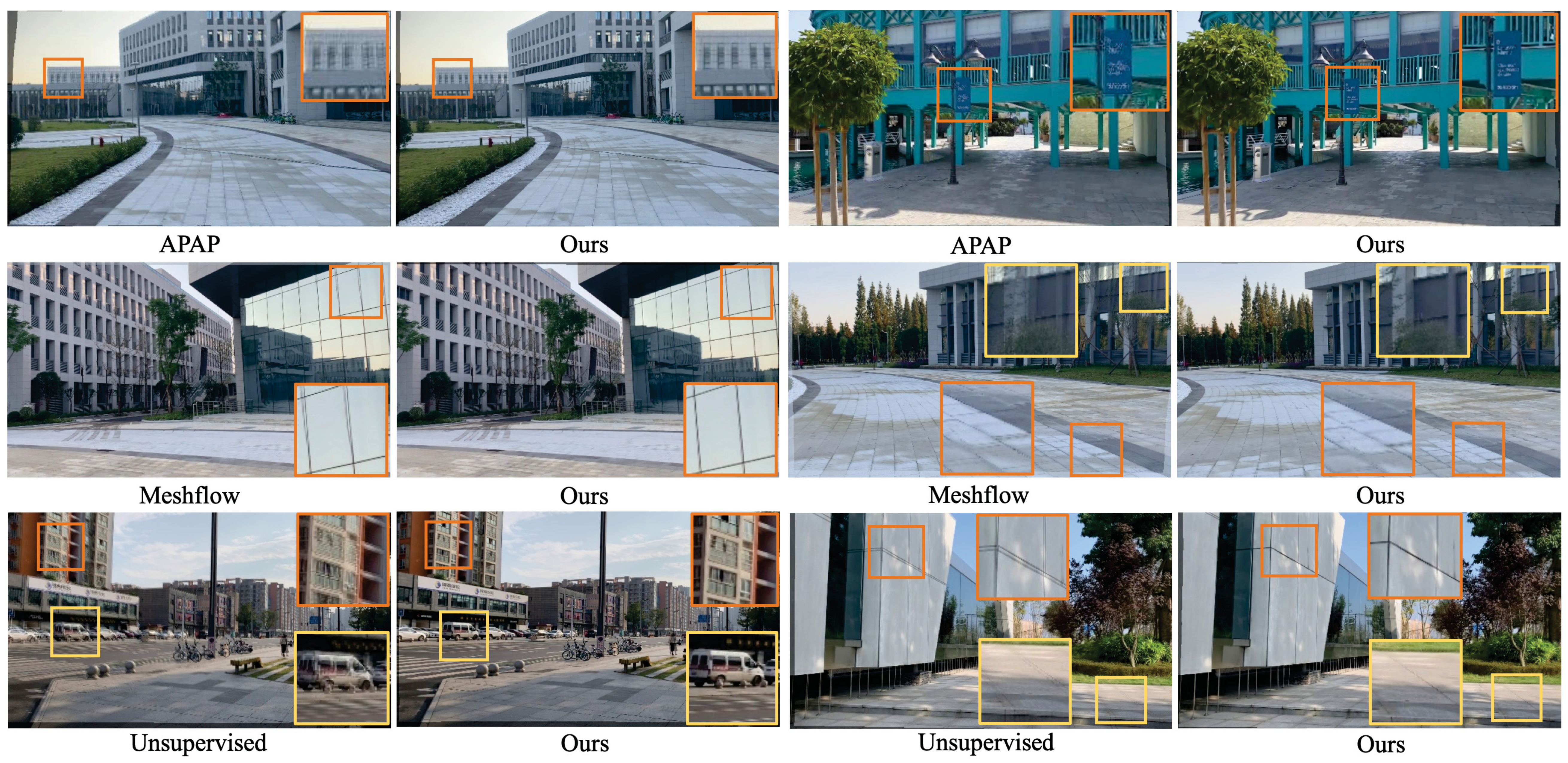}
   \caption{Comparison with existing approaches. We select three methods, APAP~\cite{zaragoza2013projective}, Meshflow~\cite{liu2016meshflow} and Unsupervised deep homography~\cite{nguyen2018unsupervised}, which are mostly related to our method for comparisons. We use zoom-in windows to highlight some regions. }\vspace{-2mm}
\label{fig:comparison}
\end{figure*}

The source image is warped to the target image, where two images are blended for illustration. Methods who produces clearer blended images indicate good alignment. For each method we show two examples as shown in Figure~\ref{fig:comparison}. The first, second, and third row shows the comparison with As-Projective-As-Possible(APAP), Meshflow, and Unsupervised deep homography approaches, respectively, in which our results are shown in the second and forth columns. We highlight some regions for clearer illustration.

\paragraph{Quantitative comparison.}
We verify the performances with our annotated points in the testing set. The comparison is based on the categories. Specifically, we use the estimated mesh/homography to transform the source points to the target points. The average $l_2$ distances are recorded as an evaluation metric. We report the performances for each category as well as the overall averaged scores in Table~\ref{tab:ablation}. Small number indicates better alignment. Table~\ref{tab:ablation} `Eye' refers to the identity matrix, indicating the original $l_2$ distances if no alignment is performed. As can be seen, the original $l_2$ distances are high, around $7\sim8$ pixels. After alignment, all methods decrease the original score, indicating that the alignment takes effect. Among all candidates, our method achieves the best result. In particular, we achieved average score of $1.99$, which surpassed the two competitors with a relatively large margin. Meshflow achieved $2.44$ and unsupervised deep homography achieved $2.45$ on average.

\subsection{Ablation Studies}
We verify the effectiveness of our design of content adaptive capability, we design two experiments, with and without mask and with fixed mesh resolutions.

\vspace{-0.2em}
\paragraph{W/o mask.}
We exclude the mask component in our pipeline to produce the result for comparisons. Table~\ref{tab:ablation} `w/o mask' shows the result. As seen, without the mask, the average performance drops from $1.99$ to $2.50$. Therefore, the mask is important during the meshflow estimation. In particular, for the low texture(LT) category, score without mask is $3.16$ while score with mask is $2.20$, improving $0.96$, which indicates that the mask is particularly helpful with respect to the LT category. For other categories, the scores with mask are also improved to a certain extent.

\paragraph{Mesh resolutions.}
We train several different fixed mesh resolutions to compare with our adaptive mesh resolution. Table~\ref{tab:ablation} shows the results. In particular, we conduct $1\times1$ mesh, $4\times4$ mesh and $16\times16$ mesh. As can be seen, non of these fixed resolutions can achieve comparable results as our adaptive mesh resolution.
We further demonstrate some visual comparisons with respect to $4\times4$ mesh and $16\times16$ mesh in Figure~\ref{fig:meshresolution}.

\section{Conclusion}\label{sec:conclu}
We have presented a network architecture for deep mesh flow estimation with content-aware abilities. Traditional feature-based methods heavily relies on the quality of image features which are vulnerable to low-texture and low-light scenes. Large foreground also causes troubles for RANSAC outlier removal. Previous deep based homography pay less attention to the depth disparity issue. They treat the image content equally which can be influenced by non-planer structures and dynamic objects. Our network learns a mask during the estimation to reject outlier regions for robust mesh flow estimation. In addition, we calculate loss with respect to our learned deep features instead of directly comparing the image contents. Moreover, we have provided a comprehensive dataset for two view alignment. The dataset have been divided into 5 categories, regular, low-texture, low-light, small-foregrounds, and large-foregrounds, to evaluate the estimation performance with respect to different aspects. The comparison with previous methods show the effectiveness of our method.

{\small
\bibliographystyle{unsrt}
\bibliography{DeepMeshflow}
}

\end{document}